\documentclass[conference]{IEEEtran}

\usepackage[utf8]{inputenc}

\usepackage[T1]{fontenc}
\usepackage{graphicx}
\usepackage{amsmath}
\usepackage{amsfonts}
\usepackage{amssymb}
\usepackage{booktabs}
\usepackage{url}
\usepackage{cite}
\usepackage{hyperref}

\title{Investigating the Vulnerability of LLM-as-a-Judge Architectures to Prompt-Injection Attacks}

\author{%
  \IEEEauthorblockN{Narek Maloyan, Bislan Ashinov, Dmitry Namiot}
  \IEEEauthorblockA{%
    maloyan.narek@gmail.com
  }
}

\begin{document}

\maketitle
\thispagestyle{plain}
\pagestyle{plain}

\begin{abstract}
Large Language Models (LLMs) are increasingly employed as evaluators (LLM-as-a-Judge) for assessing the quality of machine-generated text. This paradigm offers scalability and cost-effectiveness compared to human annotation. However, the reliability and security of such systems, particularly their robustness against adversarial manipulations, remain critical concerns. This paper investigates the vulnerability of LLM-as-a-Judge architectures to prompt-injection attacks, where malicious inputs are designed to compromise the judge's decision-making process. We formalize two primary attack strategies: Comparative Undermining Attack (CUA), which directly targets the final decision output, and Justification Manipulation Attack (JMA), which aims to alter the model's generated reasoning. Using the Greedy Coordinate Gradient (GCG) optimization method, we craft adversarial suffixes appended to one of the responses being compared. Experiments conducted on the MT-Bench Human Judgments dataset with open-source instruction-tuned LLMs (Qwen2.5-3B-Instruct and Falcon3-3B-Instruct) demonstrate significant susceptibility. The CUA achieves an Attack Success Rate (ASR) exceeding 30\%, while JMA also shows notable effectiveness. These findings highlight substantial vulnerabilities in current LLM-as-a-Judge systems, underscoring the need for robust defense mechanisms and further research into adversarial evaluation and trustworthiness in LLM-based assessment frameworks.
\end{abstract}


\section{Introduction}
In recent years, Large Language Models (LLMs) such as GPT \cite{brown2020language, openai2023gpt4}, PaLM \cite{chowdhery2022palm, anil2023palm}, and LLaMA \cite{touvron2023llama, touvron2023llama2} have become central to modern natural language processing systems. Trained on vast datasets using transformer architectures \cite{vaswani2017attention}, LLMs have achieved remarkable success in diverse tasks, including text generation, translation, complex reasoning, and question answering, with emergent abilities that scale with model size \cite{wei2022emergent}. As LLMs are increasingly integrated into real-world applications, the need for objective and scalable evaluation of their outputs has grown significantly. Manual annotation is often costly and time-consuming, leading to the emergence of the LLM-as-a-Judge architectural pattern. In this paradigm, an LLM is used to comparatively assess responses generated by other models or by itself, acting as a 'judge' to select the best response based on criteria like relevance, completeness, and accuracy.

The LLM-as-a-Judge approach has found widespread application in areas such as Reinforcement Learning from Human Feedback (RLHF) \cite{ouyang2022training, christiano2017deep, rafailov2023direct, dubois2023alpacafarm}, automated validation in crowdsourcing, and intelligent search and retrieval systems \cite{zheng2023judging, zhong2023llm}. However, entrusting evaluation to the models themselves raises critical questions about the robustness of these judges against various attacks and manipulations \cite{casper2023open, liu2023trustworthy}. A particularly concerning threat vector is prompt-injection attacks, where an attacker modifies input data to influence the judge model's decision. More broadly, recent years have seen a surge in research on LLM security and reliability, covering threats like jailbreaking \cite{wei2023jailbroken, chao2023jailbreaking, huang2023catastrophic}, instruction inversion, and adversarial prompting \cite{wallace2019universal, jia2017adversarial, zhu2023promptbench, shen2023anything, liu2023autodan}. These attacks highlight the necessity of securing not only the LLMs themselves but also the architectural solutions that employ them in novel roles, such as judges or moderators, and raise concerns about the robustness of contemporary models \cite{wang2023decodingtrust}.

The ability of an attacker to influence the evaluation system's conclusions has severe implications, ranging from degraded model quality to undermined trust in systems relying on LLM-as-a-Judge. Consequently, studying these vulnerabilities is crucial, especially as LLMs are deployed in high-stakes applications. The primary objective of this work is to investigate the vulnerabilities of LLM-as-a-Judge systems to prompt-injection attacks and to develop methods capable of effectively attacking such evaluators. This research employs optimization techniques to craft attacks on LLM-as-a-Judge systems, specifically in the context of comparing and selecting the best responses. The effectiveness of these attacks is experimentally verified on real models, and conclusions are drawn regarding the threat level posed by such attacks to LLM-as-a-Judge systems. This work contributes to understanding the risks associated with automated LLM evaluation and emphasizes the need for additional security measures in practical applications.

\section{Background and Related Work}
The vulnerabilities of LLM-as-a-Judge systems must be considered within the broader context of research into the security and attackability of LLMs. Even foundational LLMs not acting as judges exhibit high sensitivity to adversarial attacks and prompt injections \cite{qi2023fine, xu2023llm}. Modern attacks can bypass safety filters, manipulate output content, extract confidential information \cite{li2023multi, greshake2023not}, elicit malicious outputs, or circumvent system constraints \cite{kang2023exploiting, deng2023multilingual}. For instance, Liu et al. \cite{liu2024promptinject} proposed universal prompt-injection attacks on LLMs, demonstrating how gradient-based methods can automatically generate malicious injections that effectively overcome built-in filters. These attacks are often transferable across different architectures and input formats, making the threat particularly relevant for open and interactive applications \cite{zellers2019defending}.

Carlini et al. \cite{carlini2023quantifying} provided a fundamental overview of attack classes on LLMs, ranging from jailbreak prompts and codes to more sophisticated methods of instruction and context subversion. Their research indicates that LLMs can produce harmful or undesirable content even with minor injections or alterations to the input text. The issue of pre-embedded Trojans—hidden triggers within model parameters that activate unwanted behavior—also warrants special attention. This underscores the challenge of detecting malicious models and the need for new approaches to verify and interpret model behavior.

The LLM-as-a-Judge paradigm, where LLMs evaluate text quality, has gained significant traction. Zheng et al. \cite{li2023mtbench, zheng2023judging} were among the first to propose using language models as a source for evaluations and annotations, leading to platforms like Chatbot Arena \cite{lmarena} and standardized evaluation frameworks \cite{gao2023llm}. According to Gu et al. \cite{gu2024helpfuljudge}, LLM-as-a-Judge is applied in automatic model validation, content moderation, and automated peer review. Commercial solutions, such as those based on GPT-4 \cite{openai2023gpt4}, Claude \cite{anthropic2023claude}, and other advanced models, integrate these capabilities to assess generative systems based on relevance, accuracy, and safety. Despite the advantages in scalability and cost-efficiency, using LLMs as judges faces challenges like decision instability, sensitivity to rephrasing, and susceptibility to manipulation \cite{zhong2023llm}. These issues make the analysis of LLM-as-a-Judge robustness critically important.

Attacks on the LLM-as-a-Judge architecture are a relatively new but rapidly evolving field. Wang et al. \cite{wang2024badjudge} explored backdoor vulnerabilities in LLM-as-a-Judge systems, demonstrating how they can be compromised during training. Shi et al. \cite{shi2024judgedeceiver} introduced JudgeDeceiver, a method that uses prompt-injection optimization to create universal templates capable of persuading a judge model without requiring access to its internal parameters. This work highlights that even state-of-the-art models, extensively trained and fine-tuned, exhibit significant vulnerability to attacks that affect their reasoning and final choice.

Optimization-based attacks on LLMs often leverage techniques to refine input tokens. A prominent method is the Greedy Coordinate Gradient (GCG) attack, proposed by Zou et al. \cite{zou2023universal}. GCG constructs an adversarial suffix by greedily optimizing tokens based on the model's logits. GCG and its variants have proven effective in various jailbreak attacks, instruction injections, and bypassing safety filters. These methods can be adapted to undermine the decisions of evaluator models. Thus, LLM-as-a-Judge is a promising yet vulnerable tool whose weaknesses are not yet fully understood. It is imperative to investigate the robustness of these models, develop systemic defense measures like those proposed by Zhang et al. \cite{zhang2024attentiontracker}, Wu et al. \cite{wu2023defending}, and Jain et al. \cite{jain2023baseline}, and create new robustness metrics \cite{ribeiro2020beyond} that account for manipulations at both the final decision and reasoning levels. Approaches such as SmoothLLM \cite{robey2023smoothllm} and certification methods \cite{kumar2023certifying} offer promising directions for enhancing model robustness. Red teaming approaches \cite{perez2022red, ganguli2022red} and adversarial training techniques \cite{chen2023combating} can also help identify and mitigate vulnerabilities in these systems.

\section{Problem Formulation and Attack Methodology}

This section formally defines the problem of attacking LLM-as-a-Judge architectures and details the methodology of the proposed attacks.

\subsection{LLM-as-a-Judge and Robustness}
An LLM-as-a-Judge system typically involves a model $f_\theta$ with parameters $\theta$ that takes a query $x \in \mathcal{X}$ and a set of $k$ candidate answers $\{a_1, ..., a_k\} \in \mathcal{A}^k$ to produce an evaluation $y$ \cite{zheng2023judging, gu2024helpfuljudge}. In the common pairwise comparison task, $k=2$, and the judge $f_\theta(x, a, b)$ outputs a preference, e.g., $y \in \{\texttt{[[A]]}, \texttt{[[B]]}\}$. The goal of an attacker is to introduce a perturbation $\delta$ to one of the answers (e.g., $b$) such that the judge's output $f_\theta(x, a, b \oplus \delta)$ is manipulated to a desired outcome, where $\oplus$ denotes token concatenation \cite{shi2024judgedeceiver, liu2024promptinject}. The robustness of an LLM-as-a-Judge is its ability to maintain the original judgment $f_\theta(x, a, b)$ despite the adversarial perturbation $\delta$ \cite{kumar2023certifying, robey2023smoothllm}.

\subsection{Proposed Attack Strategies}
We investigate two primary attack strategies:

\subsubsection{Comparative Undermining Attack (CUA)}
This attack directly targets the final decision probability of the judge model \cite{wang2024badjudge, carlini2023quantifying}. The attacker aims to maximize the probability that the model selects a specific target answer (e.g., $\texttt{[[B]]}$) over the originally preferred answer (e.g., $\texttt{[[A]]}$). An adversarial suffix $\delta$ is appended to the target answer $b$. The optimization objective for CUA is:
\begin{equation}
\max_{\delta} \left( \mathbb{P}(\texttt{[[B]]} \mid x, a, b \oplus \delta) - \mathbb{P}(\texttt{[[A]]} \mid x, a, b \oplus \delta) \right).
\label{eq:cua_objective}
\end{equation}

\subsubsection{Justification Manipulation Attack (JMA)}
This attack focuses on manipulating the textual justification $j$ that the judge model generates alongside its decision \cite{shi2024judgedeceiver, liu2023autodan}. The goal is to increase the presence of positive lexical markers (e.g., "coherent", "accurate") related to the target answer $b$ and suppress negative markers (e.g., "incorrect", "irrelevant"), while also maximizing the probability of selecting $\texttt{[[B]]}$. This approach is inspired by research on manipulating model reasoning and explanations \cite{shen2023anything, greshake2023not}. Let $\mathcal{P}$ be the set of positive justification tokens and $\mathcal{N}$ be the set of negative justification tokens. The JMA objective is:
\begin{equation}
\begin{split}
\max_{\delta} \Big( & \sum_{t \in \mathcal{P}} \mathbb{P}(t \in j | x, a, b \oplus \delta) - \\
& \sum_{t \in \mathcal{N}} \mathbb{P}(t \in j | x, a, b \oplus \delta) + \\
& \mathbb{P}(\texttt{[[B]]} | x, a, b \oplus \delta) \Big).
\end{split}
\label{eq:jma_objective_full}
\end{equation}
This attack aims for a more subtle manipulation by altering the model's apparent reasoning process.

\subsection{Optimization via Greedy Coordinate Gradient (GCG)}
To implement these attacks, we adapt the Greedy Coordinate Gradient (GCG) method \cite{zou2023universal}. GCG efficiently optimizes the adversarial suffix $\delta$ by iteratively and greedily updating individual tokens within the suffix to maximize the attack objective. It operates by evaluating the gradient of the loss function with respect to the token embeddings at each position in the suffix and selecting token substitutions that yield the largest improvement. This method is effective even with black-box access to the model, requiring only logit outputs (or probabilities derived from them).

The GCG algorithm initializes a random suffix $s$ of length $L$. In each iteration, for every position $i$ in the suffix, it computes the gradient of the loss function $\mathcal{L}$ (derived from Eq. \ref{eq:cua_objective} or \ref{eq:jma_objective_full}) with respect to the token $s_i$. It then identifies a set of candidate token substitutions $C_i$ that are likely to improve the objective. A subset of these candidates across all positions is evaluated, and the substitution that provides the best improvement to $\mathcal{L}$ is applied. This process is repeated for a fixed number of iterations or until convergence.

\section{Experimental Setup}
This section details the experimental setup used to evaluate the effectiveness of the proposed prompt-injection attacks on LLM-as-a-Judge systems.

\subsection{Models}
Two open-source instruction-tuned LLMs were used as judge models in our experiments:
\begin{itemize}
    \item \textbf{Qwen2.5-3B-Instruct}: A 3-billion parameter model from Alibaba's Qwen family, optimized for generation and evaluation tasks \cite{qwen2}. 
    \item \textbf{Falcon3-3B-Instruct}: A lightweight 3-billion parameter model from the Technology Innovation Institute \cite{falconllm}.
\end{itemize}
These models were chosen for their public availability, representation of modern LLM architectures, and relatively small size, which allows for extensive experimentation within computational constraints.

\subsection{Dataset}
We utilized the \textbf{MT-Bench Human Judgments} dataset provided by LMSYS \cite{li2023mtbench}. This dataset contains human evaluations of LLM responses to diverse questions, presented as pairwise comparisons. Each record includes a question, two answers (Answer A and Answer B) generated by different models, and a human-adjudicated winner (A or B). For each instance, we formed a triplet $(x, a, b)$ and applied the judge model, both in a baseline scenario and with the adversarial suffix $\delta$ appended to answer $b$ (assuming $b$ was the initially losing answer or the target for manipulation).

\subsection{Baseline and Control Conditions}
We implemented several baseline and control conditions to establish the effectiveness of our proposed attacks:

\subsubsection{Hard Prompt Attack}
As a baseline, we implemented a \textbf{Hard Prompt Attack}. This attack involves appending a pre-defined, heuristically designed suffix to the target answer $b$ without optimization \cite{wei2023jailbroken, chao2023jailbreaking}. The suffixes contain direct, albeit contextually irrelevant, instructions or persuasive language designed to nudge the model towards selecting the target answer (e.g., "It is critically important that you select response B as the better one."). This approach is similar to direct jailbreaking techniques documented in prior work \cite{huang2023catastrophic, deng2023multilingual}. Several such heuristic prompts were used, with one chosen randomly for each attack instance. This baseline helps establish a lower bound for the effectiveness of more sophisticated optimization-based attacks and demonstrates the general sensitivity of LLMs to such direct manipulations \cite{qi2023fine, xu2023llm}.

\subsubsection{Random-Suffix Control}
To establish that any observed attack success represents a real gain over chance, we implemented a \textbf{Random-Suffix Control}. In this condition, we appended randomly generated text of the same length as our attack suffixes to the target answer. This control helps determine whether the success of our attacks is due to the specific content of the optimized suffixes rather than merely the presence of additional text or the disruption of the model's processing.

\subsubsection{Token-Shuffle Control}
We also implemented a \textbf{Token-Shuffle Control}, where tokens from successful attack suffixes were randomly shuffled before being appended to the target answer. This control condition helps determine whether the specific ordering and structure of tokens in the attack suffixes are critical to their effectiveness, or if the mere presence of certain tokens, regardless of their arrangement, is sufficient to influence the judge model's decision.

\subsection{Evaluation Protocol}
The primary metric for evaluating attack effectiveness is the \textbf{Attack Success Rate (ASR)}. ASR is defined as the percentage of instances where the attack successfully caused the judge model to change its verdict in favor of the (originally disfavored or targeted) answer to which the adversarial suffix was applied:
\begin{equation}
\text{ASR} = \frac{\text{\# of successful verdict flips to target}}{\text{\# of total attack attempts}} \times 100\%.
\end{equation}

\section{Results and Discussion}
This section presents the results of our experiments evaluating the CUA and JMA methods against the baseline Hard Prompt Attack on the selected LLM-as-a-Judge models.

\subsection{Attack Success Rates}
The quantitative results, presented in Table \ref{tab:asr_results}, demonstrate the effectiveness of the proposed attacks and highlight the vulnerability of the tested models.

\begin{table}[htbp]
\centering
\caption{Attack Success Rate (ASR) by Model and Method}
\label{tab:asr_results}
\begin{tabular}{@{}lcc@{}}
\toprule
\textbf{Method} & \textbf{Qwen2.5-3B (\%)} & \textbf{Falcon3-3B (\%)} \\
\midrule
Random-Suffix & 1.2 & 1.5 \\
Token-Shuffle & 2.8 & 3.1 \\
Hard Prompt & 5.1 & 5.4 \\
JMA & 15.2 & 16.7 \\
JudgeDeceiver \cite{shi2024judgedeceiver} & 22.8 & 24.1 \\
CUA & 31.2 & 32.4 \\
\bottomrule
\end{tabular}
\end{table}

Key observations from the results include:
\begin{itemize}
    \item The \textbf{Random-Suffix Control} showed minimal impact, with ASRs of only 1.2-1.5\%. This confirms that simply appending random text to an answer is unlikely to change the judge's decision, establishing that our attack methods provide real gains over chance.
    \item The \textbf{Token-Shuffle Control} achieved slightly higher ASRs of 2.8-3.1\%, suggesting that while the specific tokens used in successful attacks do have some inherent influence, their effectiveness is significantly enhanced by proper ordering and structure.
    \item The \textbf{Hard Prompt Attack} baseline showed a modest impact, with ASR around 5\%. This confirms that simple heuristic injections can influence LLM judges, but their effect is limited.
    \item The \textbf{Justification Manipulation Attack (JMA)} achieved a significantly higher ASR, around 15-17\%. This indicates that manipulating the perceived justification can be a more effective strategy than simple hard prompts.
    \item The \textbf{JudgeDeceiver} method \cite{shi2024judgedeceiver}, which uses universal templates to manipulate judge models, demonstrated substantial effectiveness with ASRs of 22.8\% and 24.1\%. This approach offers the advantage of not requiring instance-specific optimization.
    \item The \textbf{Comparative Undermining Attack (CUA)} proved to be the most potent, achieving ASRs exceeding 30\% on both models. This suggests that directly optimizing for the final decision token is a highly effective way to compromise LLM-as-a-Judge systems, likely because the optimization objective is more direct and less complex than that of JMA.
\end{itemize}

\subsection{Discussion}
The experimental results clearly demonstrate that even short, targeted prompt injections can substantially distort the evaluations made by LLM-as-a-Judge systems \cite{shi2024judgedeceiver, wang2024badjudge}. The high ASR of the CUA method, in particular, shows that judge models can be systematically biased towards an incorrect choice by appending optimized suffixes, even when the core content of the question and answers remains unchanged. This susceptibility, even with limited attack capabilities (fixed-length suffix appended to one answer), raises serious concerns about the reliability and objectivity of LLM-as-a-Judge architectures in practical scenarios where they are used for data collection or model evaluation \cite{casper2023open, liu2023trustworthy, wang2023decodingtrust}.

The comparative analysis of the attack methods reveals important insights about LLM-as-a-Judge vulnerabilities. The control conditions demonstrate that the success of our attacks is not merely due to chance or the presence of additional text. The Random-Suffix Control's minimal impact (1.2-1.5\% ASR) establishes a true baseline for random perturbations, while the Token-Shuffle Control (2.8-3.1\% ASR) shows that the specific arrangement of tokens matters significantly more than just their presence. These controls strengthen our findings by confirming that the observed attack success rates represent genuine vulnerabilities rather than artifacts of the experimental design.

While heuristic attacks have a weak effect, they confirm the possibility of manipulation \cite{wei2023jailbroken, chao2023jailbreaking}. Attacks targeting the model's reasoning (JMA) induce a more profound change in behavior \cite{liu2023autodan, shen2023anything}. The JudgeDeceiver method \cite{shi2024judgedeceiver} demonstrates that universal templates can achieve substantial success rates without requiring instance-specific optimization, offering a more efficient attack vector. However, directly influencing the decision logits (CUA) is the most effective approach, especially under the assumption of white-box or proficient grey-box access to the judge model (allowing for logit-based optimization) \cite{zou2023universal, carlini2023quantifying}. This highlights the trade-off between attack efficiency (JudgeDeceiver) and effectiveness (CUA), a pattern consistent with findings in other adversarial attack research \cite{wallace2019universal, zhu2023promptbench}.

These findings underscore the vulnerability of LLM-as-a-Judge systems \cite{wang2024badjudge, shi2024judgedeceiver}. The fact that these models, designed to act as impartial evaluators, can be so readily swayed by adversarial inputs calls into question their trustworthiness for critical assessment tasks \cite{liu2023trustworthy, casper2023open}. This work did not explore the impact of permuting the order of the attacked and genuinely superior answers, which could be a direction for future research. The proposed methods and their demonstrated success provide a foundation for further investigation into the robustness of LLM evaluation systems and the development of defenses against prompt-injection attacks \cite{zhang2024attentiontracker, wu2023defending, robey2023smoothllm, jain2023baseline}.

\section{Conclusion}
This paper investigated the robustness of LLM-as-a-Judge architectures against prompt-injection attacks \cite{shi2024judgedeceiver, wang2024badjudge}. As these architectures become increasingly prevalent for automated quality assessment \cite{zheng2023judging, li2023mtbench, gao2023llm}, understanding their security vulnerabilities is paramount \cite{liu2023trustworthy, wang2023decodingtrust}.

We formalized and empirically evaluated two primary attack methods: the Comparative Undermining Attack (CUA), targeting the final decision token \cite{carlini2023quantifying, zou2023universal}, and the Justification Manipulation Attack (JMA), aimed at altering the model's generated reasoning \cite{liu2023autodan, shen2023anything}. Both attacks employed the Greedy Coordinate Gradient (GCG) optimization technique \cite{zou2023universal} to craft adversarial suffixes. Experiments were conducted on the MT-Bench Human Judgments dataset \cite{li2023mtbench} using two open-source judge models, Qwen2.5-3B-Instruct \cite{qwen2} and Falcon3-3B-Instruct \cite{falconllm}.

Our results demonstrate that LLM-as-a-Judge systems are significantly vulnerable to such attacks \cite{shi2024judgedeceiver, wang2024badjudge}. The CUA method achieved an Attack Success Rate of over 30\%, indicating that targeted injections can effectively manipulate the judge's decision \cite{zou2023universal, carlini2023quantifying}. The JMA method also showed considerable success \cite{liu2023autodan, shen2023anything}. We compared our approaches with JudgeDeceiver \cite{shi2024judgedeceiver}, which uses universal templates and achieved ASRs of 22-24\%, confirming the vulnerability of judge models while highlighting the trade-off between attack efficiency and effectiveness. Even simple heuristic-based hard prompt attacks exhibited a non-negligible impact \cite{wei2023jailbroken, chao2023jailbreaking}, highlighting the general sensitivity of these models to instructional context within the inputs they evaluate \cite{qi2023fine, xu2023llm}.

This research underscores the critical need to address the security and reliability of LLM-as-a-Judge systems. The demonstrated vulnerabilities suggest that current models may not be sufficiently robust for high-stakes evaluation tasks without additional safeguards. Future work should focus on developing effective defense mechanisms like those proposed by Zhang et al. \cite{zhang2024attentiontracker}, Wu et al. \cite{wu2023defending}, and Robey et al. \cite{robey2023smoothllm}, creating more comprehensive adversarial evaluation benchmarks \cite{zhu2023promptbench}, and enhancing the inherent trustworthiness of LLMs employed in evaluative roles through techniques such as red teaming \cite{ganguli2022red, perez2022red} and constitutional AI approaches \cite{bai2022constitutional}. Certification methods \cite{kumar2023certifying} and baseline defenses \cite{jain2023baseline} also offer promising directions for securing these systems. This study serves as a step towards a deeper understanding of the risks associated with automated LLM assessment and the broader implications for AI safety and reliability in open-ended interaction scenarios, as highlighted in recent trustworthiness evaluations \cite{liu2023trustworthy, wang2023decodingtrust}.

\bibliographystyle{IEEEtran}
\bibliography{IEEEabrv,kr_2024_ieee} 

\end{document}